\documentclass[10pt,twocolumn,letterpaper]{article}

\usepackage{cvpr}
\usepackage{times}
\usepackage{epsfig}
\usepackage{graphicx}
\usepackage{amsmath}
\usepackage{amssymb}

\usepackage{nicefrac}       % compact symbols for 1/2, etc.
\usepackage{microtype}      % microtypography

\usepackage{multirow}
\usepackage{verbatim}
\usepackage{color}
\usepackage{float}
\usepackage{enumitem}
\usepackage{booktabs}
\usepackage{tabulary,multirow,overpic,xcolor}
\usepackage[caption=false]{subfig}

\usepackage[british,UKenglish,USenglish,english,american]{babel}

\newcommand{\ve}[1]{\mathbf{#1}} % for displaying a vector
\newcolumntype{x}[1]{>{\centering\arraybackslash}p{#1pt}}

\newcommand{\bd}[1]{\textbf{#1}}
\newcommand{\app}{\raise.17ex\hbox{$\scriptstyle\sim$}}

\def\x{$\times$}
\newcolumntype{x}[1]{>{\centering\arraybackslash}p{#1pt}}

\newlength\savewidth\newcommand\shline{\noalign{\global\savewidth\arrayrulewidth
  \global\arrayrulewidth 1pt}\hline\noalign{\global\arrayrulewidth\savewidth}}
\newcommand{\tablestyle}[2]{\setlength{\tabcolsep}{#1}\renewcommand{\arraystretch}{#2}\centering\footnotesize}

\makeatletter\renewcommand\paragraph{\@startsection{paragraph}{4}{\z@}
  {.5em \@plus1ex \@minus.2ex}{-.5em}{\normalfont\normalsize\bfseries}}\makeatother

\definecolor{citecolor}{RGB}{34,139,34}
\usepackage[pagebackref=true,breaklinks=true,letterpaper=true,colorlinks,
  citecolor=citecolor,bookmarks=false]{hyperref}

\cvprfinalcopy % *** Uncomment this line for the final submission

\def\cvprPaperID{***} % *** Enter the CVPR Paper ID here

\ifcvprfinal\fi
\begin{document}

%%%%%%%%% TITLE
\title{Non-local Neural Networks \vspace{-.5em}}

\author{
Xiaolong Wang\textsuperscript{1,2}\footnotemark \qquad
Ross Girshick\textsuperscript{2} \qquad
Abhinav Gupta\textsuperscript{1} \qquad
Kaiming He\textsuperscript{2} \vspace{.3em}\\
\textsuperscript{1}Carnegie Mellon University \qquad\qquad \textsuperscript{2}Facebook AI Research
\vspace{-.6em}
}

\maketitle
\renewcommand*{\thefootnote}{\fnsymbol{footnote}}
\setcounter{footnote}{1}
\footnotetext{Work done during an internship at Facebook AI Research.}
\renewcommand*{\thefootnote}{\arabic{footnote}}
\setcounter{footnote}{0}

%\thispagestyle{empty}

%%%%%%%%% ABSTRACT
\begin{abstract}
\vspace{-.5em}
Both convolutional and recurrent operations are building blocks that process one local neighborhood at a time. In this paper, we present non-local operations as a generic family of building blocks for capturing long-range dependencies. Inspired by the classical non-local means method \cite{Buades2005} in computer vision, our non-local operation computes the response at a position as a weighted sum of the features at all positions. This building block can be plugged into many computer vision architectures. On the task of video classification, even without any bells and whistles, our non-local models can compete or outperform current competition winners on both Kinetics and Charades datasets.
In static image recognition, our non-local models improve object detection/segmentation and pose estimation on the COCO suite of tasks. Code is available at \url{https://github.com/facebookresearch/video-nonlocal-net}.
\end{abstract}
\vspace{-1em}
%%%%%%%%% BODY TEXT
\section{Introduction}

Capturing \emph{long-range} dependencies is of central importance in deep neural networks. For sequential data (\eg, in speech, language), \emph{recurrent} operations \cite{Rumelhart1986,Hochreiter1997} are the dominant solution to long-range dependency modeling. For image data, long-distance dependencies are modeled by the large receptive fields formed by deep stacks of \emph{convolutional} operations \cite{Fukushima1982,LeCun1989}.

Convolutional and recurrent operations both process a \emph{local} neighborhood, either in space or time; thus long-range dependencies can only be captured when these operations are applied repeatedly, propagating signals progressively through the data. Repeating local operations has several limitations. First, it is computationally inefficient. Second, it causes optimization difficulties that need to be carefully addressed \cite{Hochreiter1997,He2016}. Finally, these challenges make multi-hop dependency modeling, \eg, when messages need to be delivered back and forth between distant positions, difficult.

\begin{figure}[t]
\centering
\includegraphics[width=1\linewidth]{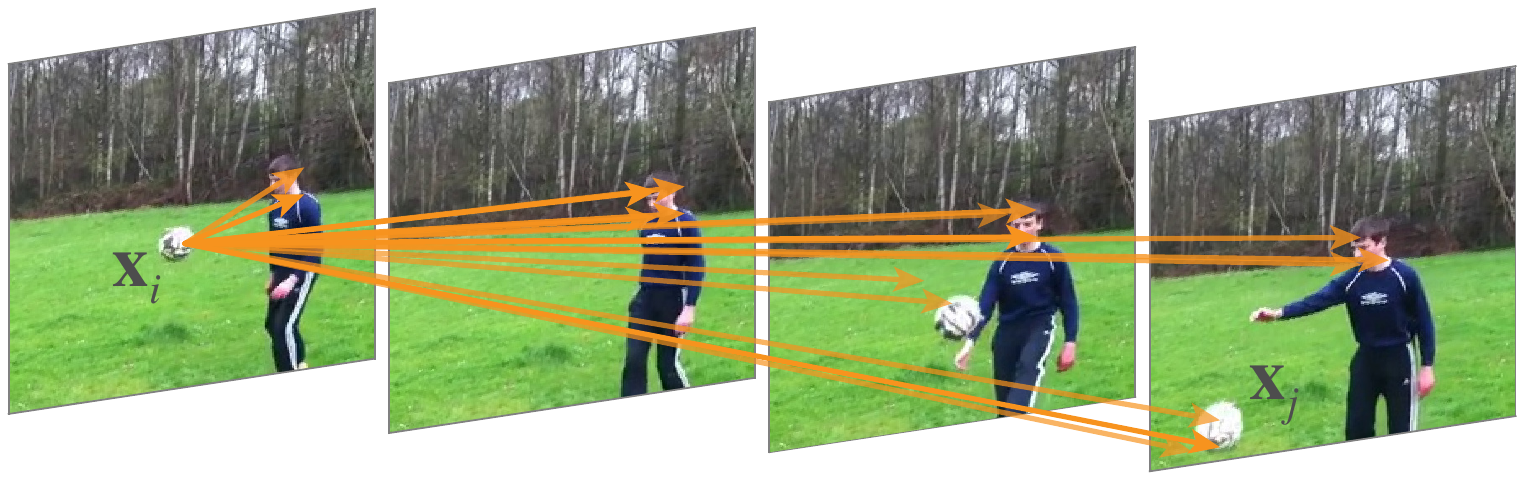}
\vspace{-1em}
\caption{A spacetime \emph{\bd{non-local}} operation in our network trained for video classification in Kinetics. A position $\ve{x}_i$'s response is computed by the weighted average of the features of \emph{all} positions $\ve{x}_j$ (only the highest weighted ones are shown here). In this example computed by our model, note how it relates the ball in the first frame to the ball in the last two frames. More examples are in Figure~\ref{fig:examples}.}
\label{fig:teaser}
\vspace{-1.1em}
\end{figure}

In this paper, we present \emph{non-local} operations as an efficient, simple, and generic component for capturing long-range dependencies with deep neural networks. Our proposed non-local operation is a generalization of the classical non-local mean operation \cite{Buades2005} in computer vision. Intuitively, a non-local operation computes the response at a position as a weighted sum of the features at \emph{all positions} in the input feature maps (Figure~\ref{fig:teaser}). The set of positions can be in space, time, or spacetime, implying that our operations are applicable for image, sequence, and video problems.

There are several advantages of using non-local operations: (a) In contrast to the progressive behavior of recurrent and convolutional operations, non-local operations capture long-range dependencies directly by computing interactions between any two positions, regardless of their positional distance; (b) As we show in experiments, non-local operations are efficient and achieve their best results even with only a few layers (\eg, 5); (c) Finally, our non-local operations maintain the variable input sizes and can be easily combined with other operations (\eg, convolutions as we will use).

%##################################################################################################
%##################################################################################################

We showcase the effectiveness of non-local operations in the application of video classification. In videos, long-range interactions occur between distant pixels in space as well as time. A single non-local block, which is our basic unit, can directly capture these spacetime dependencies in a feedforward fashion. With a few non-local blocks, our architecures called \emph{non-local neural networks} are more accurate for video classification than 2D and 3D convolutional networks \cite{Tran2015} (including the inflated variant \cite{Carreira2017}). In addition, non-local neural networks are more computationally economical than their 3D convolutional counterparts. Comprehensive ablation studies are presented on the Kinetics \cite{Kay2017} and Charades \cite{Sigurdsson2016} datasets. \emph{Using RGB only and without any bells and whistles} (\eg, optical flow, multi-scale testing), our method achieves results on par with or better than the latest competitions winners on both datasets.

To demonstrate the generality of non-local operations, we further present object detection/segmentation and pose estimation experiments on the COCO dataset \cite{Lin2014}. On top of the strong Mask R-CNN baseline \cite{He2017}, our non-local blocks can increase accuracy on all three tasks at a small extra computational cost.
Together with the evidence on videos, these image experiments show that non-local operations are generally useful and can become a basic building block in designing deep neural networks.

\vspace{-.2em}
\section{Related Work}
\vspace{-.2em}

\paragraph{Non-local image processing.} Non-local means \cite{Buades2005} is a classical filtering algorithm that computes a weighted mean of all pixels in an image. It allows distant pixels to contribute to the filtered response at a location based on patch appearance similarity. This non-local filtering idea was later developed into BM3D (block-matching 3D) \cite{Dabov2007}, which performs filtering on a group of similar, but non-local, patches. BM3D is a solid image denoising baseline even compared with deep neural networks \cite{Burger2012}. Block matching was used with neural networks for image denoising \cite{Burger2012a,Lefkimmiatis2016}.
Non-local matching is also the essence of successful texture synthesis \cite{Efros1999}, super-resolution \cite{Glasner2009}, and inpainting \cite{Barnes2009} algorithms. 

\vspace{-.2em}
\paragraph{Graphical models.} Long-range dependencies can be modeled by graphical models such as conditional random fields (CRF) \cite{Lafferty2001,Kraehenbuehl2011}. In the context of deep neural networks, a CRF can be exploited to post-process semantic segmentation predictions of a network \cite{Chen2014}. The iterative mean-field inference of CRF can be turned into a recurrent network and trained \cite{Zheng2015,Schwing2015,Chandra2017,Harley2017,Liu2017}. 
In contrast, our method is a simpler feedforward block for computing non-local filtering. 
Unlike these methods that were developed for segmentation, our general-purpose component is applied for classification and detection. These methods and ours are also related to a more abstract model called graph neural networks \cite{Scarselli2009}.

\vspace{-.2em}
\paragraph{Feedforward modeling for sequences.} Recently there emerged a trend of using feedforward (\ie, non-recurrent) networks for modeling sequences in speech and language \cite{Oord2016,Xiong2016,Gehring2017}. In these methods, long-term dependencies are captured by the large receptive fields contributed by very deep 1-D convolutions. These feedforward models are amenable to parallelized implementations and can be more efficient than widely used recurrent models.

\vspace{-.2em}
\paragraph{Self-attention.} Our work is related to the recent \emph{self-attention} \cite{Vaswani2017} method for machine translation. A self-attention module computes the response at a position in a sequence (\eg, a sentence) by attending to all positions and taking their weighted average in an embedding space. As we will discuss in the next, self-attention can be viewed as a form of the non-local mean \cite{Buades2005}, and in this sense our work bridges self-attention for machine translation to the more general class of non-local filtering operations that are applicable to image and video problems in computer vision.

\paragraph{Interaction networks.} \emph{Interaction Networks} (IN) \cite{Battaglia2016,Watters2017} were proposed recently for modeling physical systems. They operate on graphs of objects involved in pairwise interactions. Hoshen \cite{Hoshen2017} presented the more efficient Vertex Attention IN (VAIN) in the context of multi-agent predictive modeling. Another variant, named Relation Networks \cite{Santoro2017}, computes a function on the feature embeddings at all pairs of positions in its input. Our method also processes all pairs, as we will explain ($f(\ve{x}_i, \ve{x}_j)$ in Eq.(\ref{eq:nonlocal})). While our non-local networks are connected to these approaches, our experiments indicate that the \emph{non-locality} of the model, which is orthogonal to the ideas of attention/interaction/relation (\eg, a network can attend to a local region), is the key to their empirical success. Non-local modeling, a long-time crucial element of image processing (\eg, \cite{Efros1999,Buades2005}), has been largely overlooked in recent neural networks for computer vision.

\paragraph{Video classification architectures.} A natural solution to video classification is to combine the success of CNNs for images and RNNs for sequences \cite{Yue-HeiNg2015,Donahue2015}. In contrast, feedforward models are achieved by 3D convolutions (C3D) \cite{Ji2010,Tran2015} in spacetime, and the 3D filters can be formed by ``inflating'' \cite{Feichtenhofer2016,Carreira2017} pre-trained 2D filters.
In addition to end-to-end modeling on raw video inputs, it has been found that optical flow \cite{Simonyan2014} and trajectories \cite{Wang2013a,Wang2015} can be helpful. Both flow and trajectories are off-the-shelf modules that may find long-range, non-local dependency. A systematic comparison of video architectures can be found in \cite{Carreira2017}.

\vspace{-.2em}
\section{Non-local Neural Networks}
We first give a general definition of non-local operations and then we provide several specific instantiations of it.

\subsection{Formulation}
Following the non-local mean operation \cite{Buades2005}, we define a generic non-local operation in deep neural networks as:
\begin{equation}\label{eq:nonlocal}
\ve{y}_i = \frac{1}{\mathcal{C(\ve{x})}} \sum_{\forall j}f(\ve{x}_i, \ve{x}_j)g(\ve{x}_j).
\end{equation}
Here $i$ is the index of an output position (in space, time, or spacetime) whose response is to be computed and $j$ is the index that enumerates all possible positions. $\ve{x}$ is the input signal (image, sequence, video; often their features) and $\ve{y}$ is the output signal of the same size as $\ve{x}$. A pairwise function $f$ computes a scalar (representing relationship such as affinity) between $i$ and all $j$. The unary function $g$ computes a representation of the input signal at the position $j$. The response is normalized by a factor $\mathcal{C}(\ve{x})$.

The non-local behavior in Eq.(\ref{eq:nonlocal}) is due to the fact that all positions ($\forall j$) are considered in the operation. As a comparison, a convolutional operation sums up the  weighted input in a \emph{local} neighborhood (\eg, $i-1\leq j \leq i+1$ in a 1D case with kernel size 3), and a recurrent operation at time $i$ is often based only on the current and the latest time steps (\eg, $j=i$ or $i-1$).

The non-local operation is also different from a fully-connected (\emph{fc}) layer. Eq.(\ref{eq:nonlocal}) computes responses based on relationships between different locations, whereas \emph{fc} uses learned weights. In other words, the relationship between $\ve{x}_j$ and $\ve{x}_i$ is not a function of the input data in \emph{fc}, unlike in non-local layers. Furthermore, our formulation in Eq.(\ref{eq:nonlocal}) supports inputs of \emph{variable} sizes, and maintains the corresponding size in the output.
On the contrary, an \emph{fc} layer requires a fixed-size input/output and loses positional correspondence (\eg, that from $\ve{x}_i$ to $\ve{y}_i$ at the position $i$).

A non-local operation is a flexible building block and can be easily used together with convolutional/recurrent layers. It can be added into the earlier part of deep neural networks, unlike \emph{fc} layers that are often used in the end.
This allows us to build a richer hierarchy that combines both non-local and local information.

\subsection{Instantiations}\label{sec:instantiations}
\vspace{-.2em}
Next we describe several versions of $f$ and $g$. Interestingly, we will show by experiments (Table~\ref{tab:ablation:instantiations}) that our non-local models are not sensitive to these choices, indicating that the generic non-local behavior is the main reason for the observed improvements.

For simplicity, we only consider $g$ in the form of a linear embedding: $g(\ve{x}_j)=W_g\ve{x}_j$, where $W_g$ is a weight matrix to be learned. This is implemented as, \eg, 1\x1 convolution in space or 1\x1\x1 convolution in spacetime.

Next we discuss choices for the pairwise function $f$.

\paragraph{Gaussian.} Following the non-local mean \cite{Buades2005} and bilateral filters \cite{Tomasi1998}, a natural choice of $f$ is the Gaussian function. In this paper we consider:
\begin{equation}\label{eq:gaussian}
f(\ve{x}_i, \ve{x}_j) = e^{\ve{x}_i^T\ve{x}_j}.
\end{equation}
Here $\ve{x}_i^T\ve{x}_j$ is dot-product similarity. Euclidean distance as used in \cite{Buades2005,Tomasi1998} is also applicable, but dot product is more implementation-friendly in modern deep learning platforms. The normalization factor is set as $\mathcal{C}(\ve{x})=\sum_{\forall j}f(\ve{x}_i, \ve{x}_j)$.

\paragraph{Embedded Gaussian.} A simple extension of the Gaussian function is to compute similarity in an embedding space. In this paper we consider:
\begin{equation}\label{eq:gaussian_emb}
f(\ve{x}_i, \ve{x}_j) = e^{\theta(\ve{x}_i)^T\phi(\ve{x}_j)}.
\end{equation}
Here $\theta(\ve{x}_i)=W_\theta\ve{x}_i$ and $\phi(\ve{x}_j)=W_\phi\ve{x}_j$ are two embeddings. As above, we set $\mathcal{C}(\ve{x})=\sum_{\forall j}f(\ve{x}_i, \ve{x}_j)$.

We note that \emph{the self-attention module \cite{Vaswani2017} recently presented for machine translation is a special case of non-local operations in the embedded Gaussian version.}
This can be seen from the fact that for a given $i$, $\frac{1}{\mathcal{C}(\ve{x})}f(\ve{x}_i, \ve{x}_j)$ becomes the \emph{softmax} computation along the dimension $j$. So we have $\ve{y}=\text{\emph{softmax}}(\ve{x}^TW^T_\theta W_\phi\ve{x})g(\ve{x})$, which is the self-attention form in \cite{Vaswani2017}.
As such, our work provides insight by relating this recent self-attention model to the classic computer vision method of non-local means \cite{Buades2005}, and  extends the sequential self-attention network in \cite{Vaswani2017} to a generic space/spacetime non-local network for image/video recognition in computer vision.

Despite the relation to \cite{Vaswani2017}, we show that the attentional behavior (due to softmax) is \emph{not} essential in the applications we study. To show this, we describe two alternative versions of non-local operations next.

%##################################################################################################
\begin{figure}[t]
\centering
\includegraphics[width=.75\linewidth]{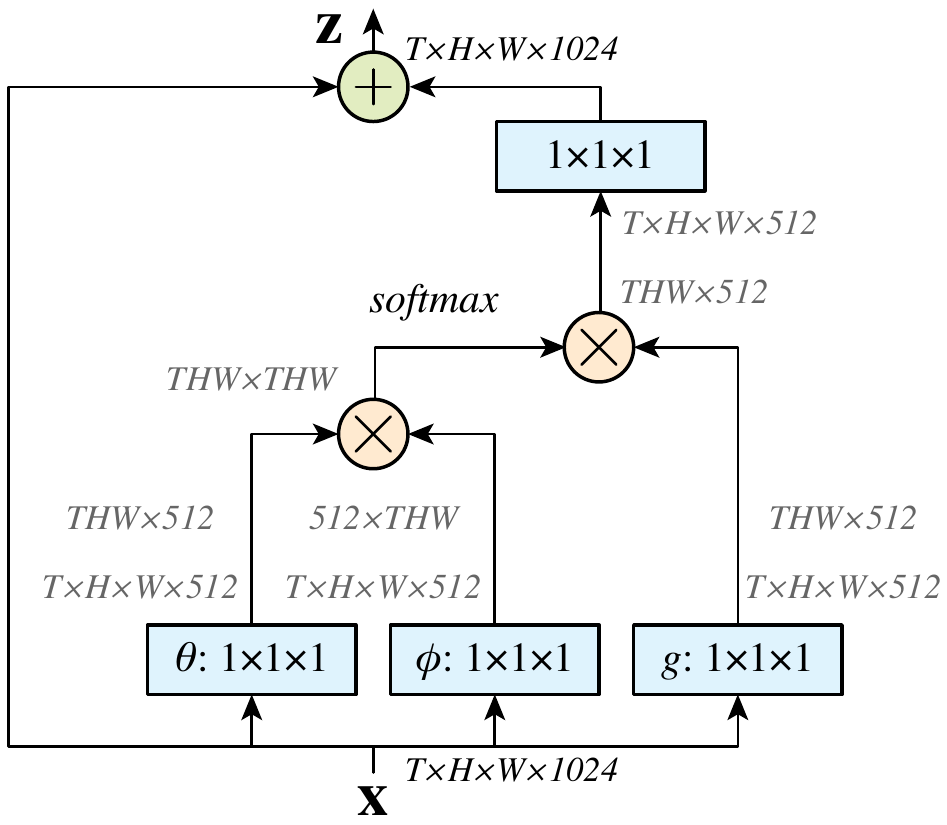}
\caption{A spacetime \textbf{non-local block}. The feature maps are shown as the shape of their tensors, \eg, $T$\x$H$\x$W$\x$1024$ for 1024 channels (proper reshaping is performed when noted). ``$\otimes$'' denotes matrix multiplication, and ``$\oplus$'' denotes element-wise sum. The softmax operation is performed on each row. The blue boxes denote 1\x1\x1 convolutions. Here we show the embedded Gaussian version, with a bottleneck of 512 channels. The vanilla Gaussian version can be done by removing $\theta$ and $\phi$, and the dot-product version can be done by replacing softmax with scaling by $1/N$.}
\label{fig:block}
\end{figure}
%##################################################################################################

\paragraph{Dot product.} $f$ can be defined as a dot-product similarity:
\begin{equation}\label{eq:dotprod}
f(\ve{x}_i, \ve{x}_j) = \theta(\ve{x}_i)^T\phi(\ve{x}_j).
\end{equation}
Here we adopt the embedded version. In this case, we set the normalization factor as $\mathcal{C}(\ve{x})=N$, where $N$ is the number of positions in $\ve{x}$, rather than the sum of $f$, because it simplifies gradient computation. A normalization like this is necessary because the input can have variable size.

The main difference between the dot product and embedded Gaussian versions is the presence of softmax, which plays the role of an activation function.

\paragraph{Concatenation.} Concatenation is used by the pairwise function in Relation Networks \cite{Santoro2017} for visual reasoning. We also evaluate a concatenation form of $f$:
\begin{equation}\label{eq:concat}
f(\ve{x}_i, \ve{x}_j) = \text{{ReLU}}(\ve{w}^T_f[\theta(\ve{x}_i),~\phi(\ve{x}_j)]).
\end{equation}
Here $[\cdot, \cdot]$ denotes concatenation and $\ve{w}_f$ is a weight vector that projects the concatenated vector to a scalar. As above, we set $\mathcal{C}(\ve{x})=N$. In this case, we adopt ReLU \cite{Nair2010} in $f$.

\vspace{1em} The above several variants demonstrate the flexibility of our generic non-local operation. We believe alternative versions are possible and may improve results.

\subsection{Non-local Block}

We wrap the non-local operation in Eq.(\ref{eq:nonlocal}) into a non-local block that can be incorporated into many existing architectures. We define a non-local block as:
\begin{equation}\label{eq:block}
\ve{z}_i = W_z \ve{y}_i + \ve{x}_i,
\end{equation}
where $\ve{y}_i$ is given in Eq.(\ref{eq:nonlocal}) and ``$+ \ve{x}_i$'' denotes a residual connection \cite{He2016}. The residual connection allows us to insert a new non-local block into any pre-trained model, without breaking its initial behavior (\eg, if $W_z$ is initialized as zero).
An example non-local block is illustrated in Figure~\ref{fig:block}. The pairwise computation in Eq.(\ref{eq:gaussian}), (\ref{eq:gaussian_emb}), or (\ref{eq:dotprod}) can be simply done by matrix multiplication as shown in Figure~\ref{fig:block}; the concatenation version in (\ref{eq:concat}) is straightforward.

The pairwise computation of a non-local block is lightweight when it is used in high-level, sub-sampled feature maps. For example, typical values in Figure~\ref{fig:block} are $T=4$, $H=W=14$ or $7$. The pairwise computation as done by matrix multiplication is comparable to a typical convolutional layer in standard networks.
We further adopt the following implementations that make it more efficient.

\paragraph{Implementation of Non-local Blocks.}
We set the number of channels represented by $W_g$, $W_\theta$, and $W_\phi$ to be half of the number of channels in $\ve{x}$. This follows the bottleneck design of \cite{He2016} and reduces the computation of a block by about a half. The weight matrix $W_z$ in Eq.(\ref{eq:block}) computes a position-wise embedding on $\ve{y}_i$, matching the number of channels to that of $\ve{x}$. See Figure~\ref{fig:block}.

A subsampling trick can be used to further reduce computation. We modify Eq.(\ref{eq:nonlocal}) as:  $\ve{y}_i = \frac{1}{\mathcal{C}(\ve{\hat{x}})} \sum_{\forall j}f(\ve{x}_i, \ve{\hat{x}}_j)g(\ve{\hat{x}}_j)$, where $\ve{\hat{x}}$ is a subsampled version of $\ve{x}$ (\eg, by pooling). We perform this in the spatial domain, which can reduce the amount of pairwise computation by 1/4. This trick does not alter the non-local behavior, but only makes the computation sparser. This can be done by adding a max pooling layer after $\phi$ and $g$ in Figure~\ref{fig:block}.

We use these efficient modifications for all non-local blocks studied in this paper.

%##################################################################################################
\newcommand{\blockb}[3]{\multirow{3}{*}{\(\left[\begin{array}{c}\text{1\x1, #2}\\[-.1em] \text{3\x3, #2}\\[-.1em] \text{1\x1, #1}\end{array}\right]\)\x#3}}
\begin{table}[t]
\footnotesize
\centering
\resizebox{0.7\columnwidth}{!}{
\tablestyle{6pt}{1.08}
\begin{tabular}{c|c|c}
\multicolumn{2}{c|}{layer} & output size \\
\shline
conv$_1$ & \multicolumn{1}{c|}{7\x7, 64, stride 2, 2, 2} & 16\x112\x112 \\
\hline
pool$_1$  & \multicolumn{1}{c|}{3\x3\x3 max, stride 2, 2, 2} & 8\x56\x56 \\
\hline
\multirow{3}{*}{res$_2$} & \blockb{256}{64}{3} & \multirow{3}{*}{8\x56\x56} \\
  &  & \\
  &  & \\
\hline
pool$_2$  & \multicolumn{1}{c|}{3\x1\x1 max, stride 2, 1, 1} & 4\x56\x56 \\
\hline
\multirow{3}{*}{res$_3$} & \blockb{512}{128}{4} & \multirow{3}{*}{4\x28\x28} \\
  &  & \\
  &  & \\
\hline
\multirow{3}{*}{res$_4$} & \blockb{1024}{256}{6} & \multirow{3}{*}{4\x14\x14}  \\
  &  & \\
  &  & \\
\hline
\multirow{3}{*}{res$_5$} & \blockb{2048}{512}{3} & \multirow{3}{*}{4\x7\x7} \\
  &  & \\
  &  & \\
\hline
\multicolumn{2}{c|}{global average pool, fc} & 1\x1\x1  \\
\end{tabular}}
\vspace{.5em}
\caption{Our \emph{baseline} ResNet-50 C2D model for video. The dimensions of 3D output maps and filter kernels are in T\x H\x W (2D kernels in H\x W), with the number of channels following.
The input is 32\x224\x224. Residual blocks are shown in brackets. % with the numbers of blocks stacked.
}
\vspace{-1em}
\label{tab:arch}
\end{table}
%##################################################################################################

%##################################################################################################
\newcommand{\sz}{.48}
\begin{figure*}[t]
\centering
\includegraphics[width=\sz\linewidth]{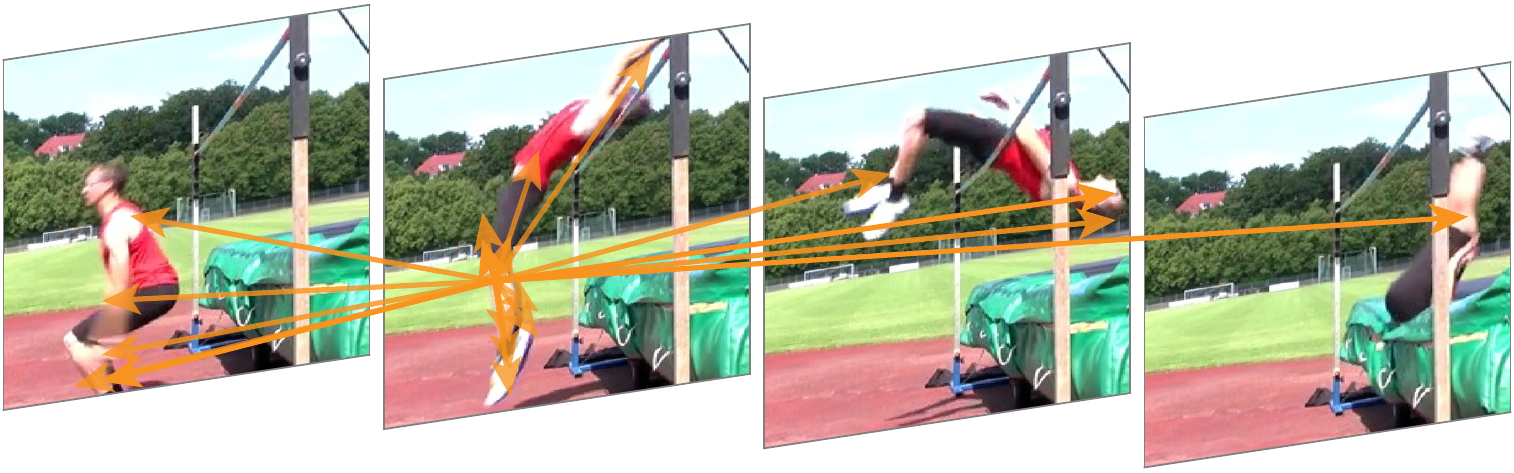}\quad
\includegraphics[width=\sz\linewidth]{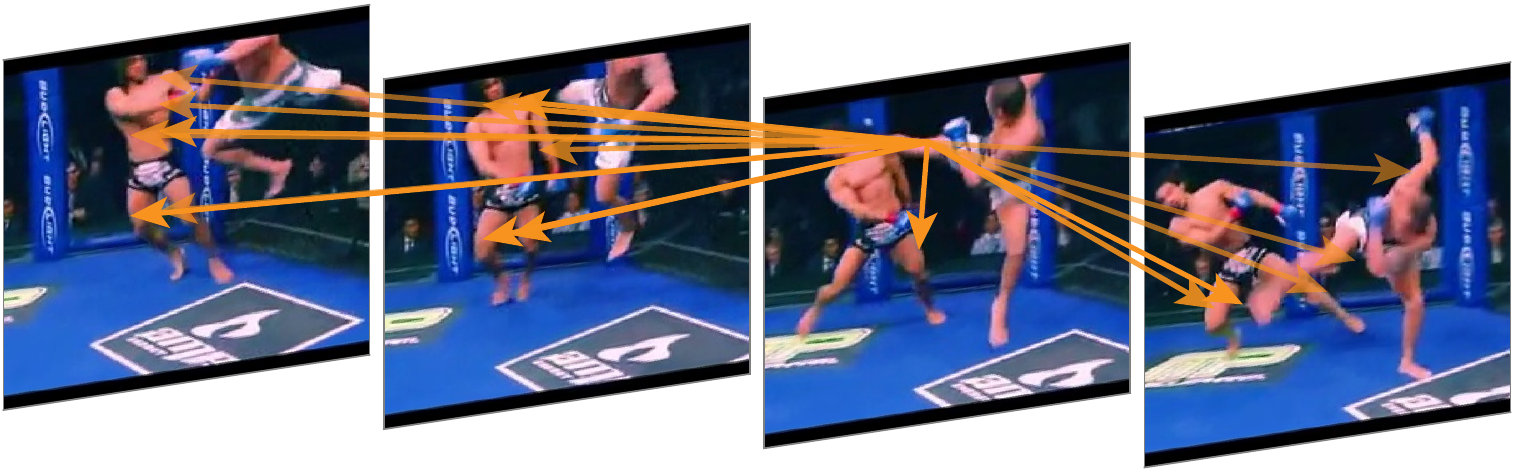}
\includegraphics[width=\sz\linewidth]{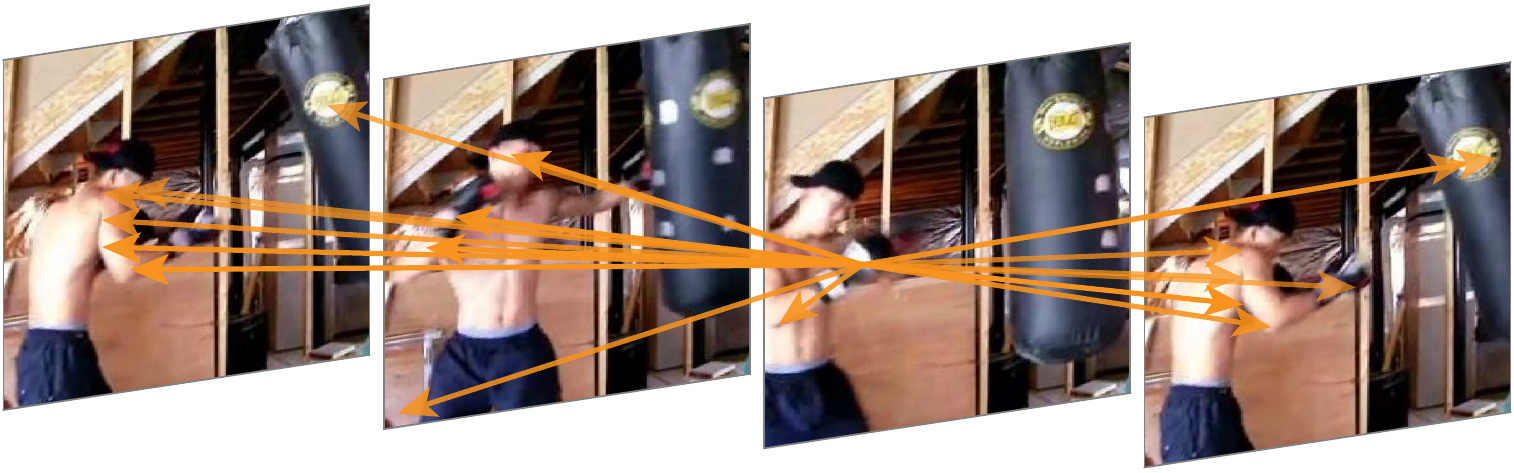}\quad
\includegraphics[width=\sz\linewidth]{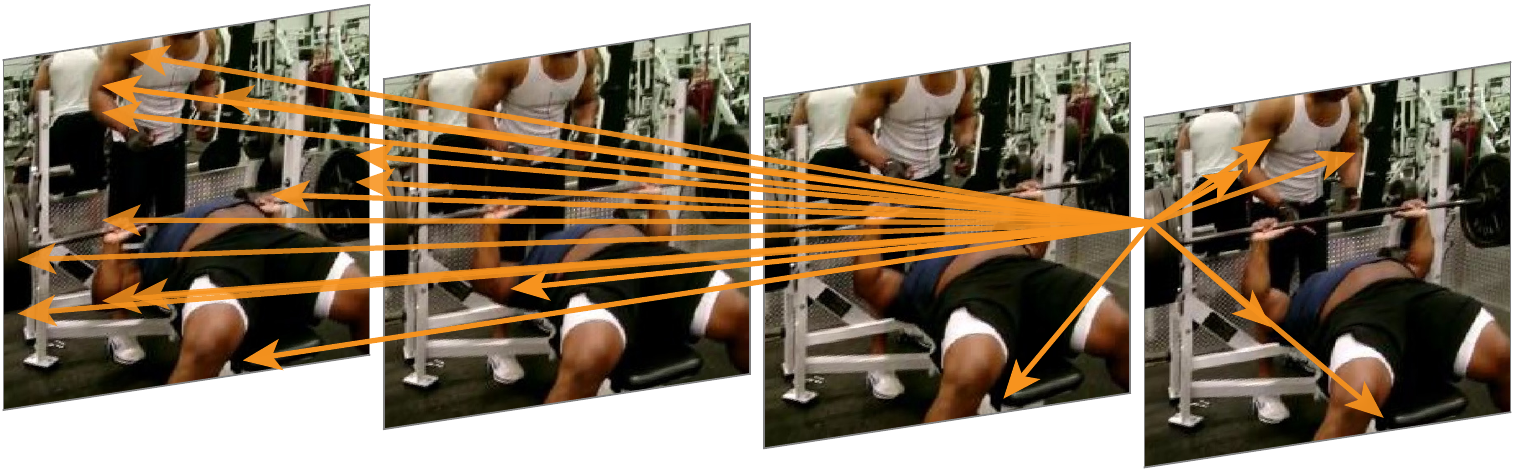}
\includegraphics[width=\sz\linewidth]{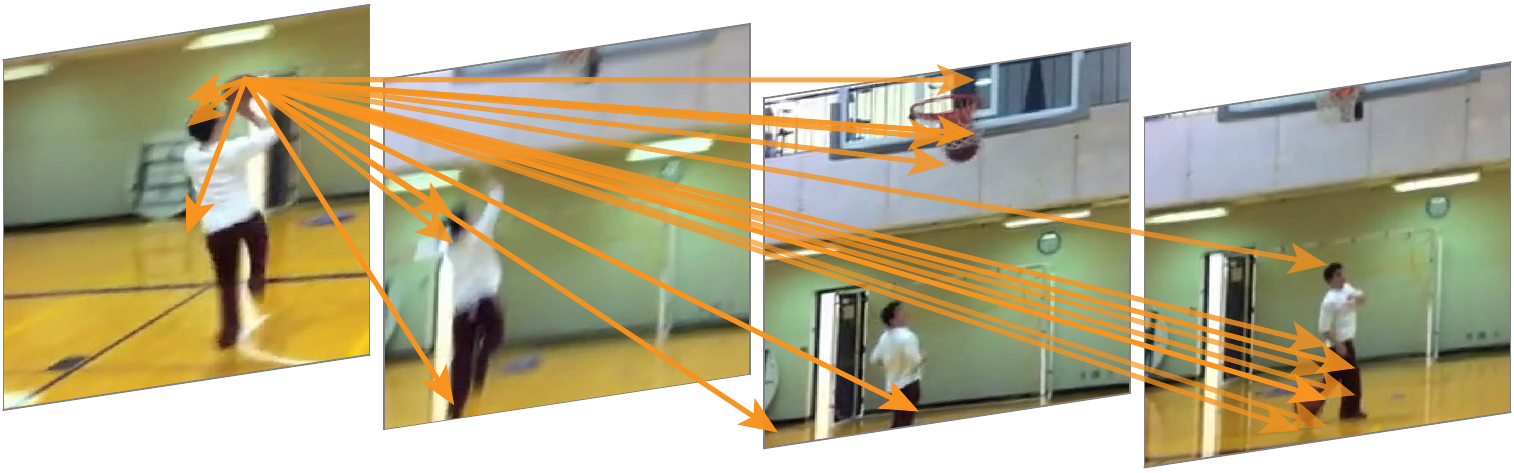}\quad
\includegraphics[width=\sz\linewidth]{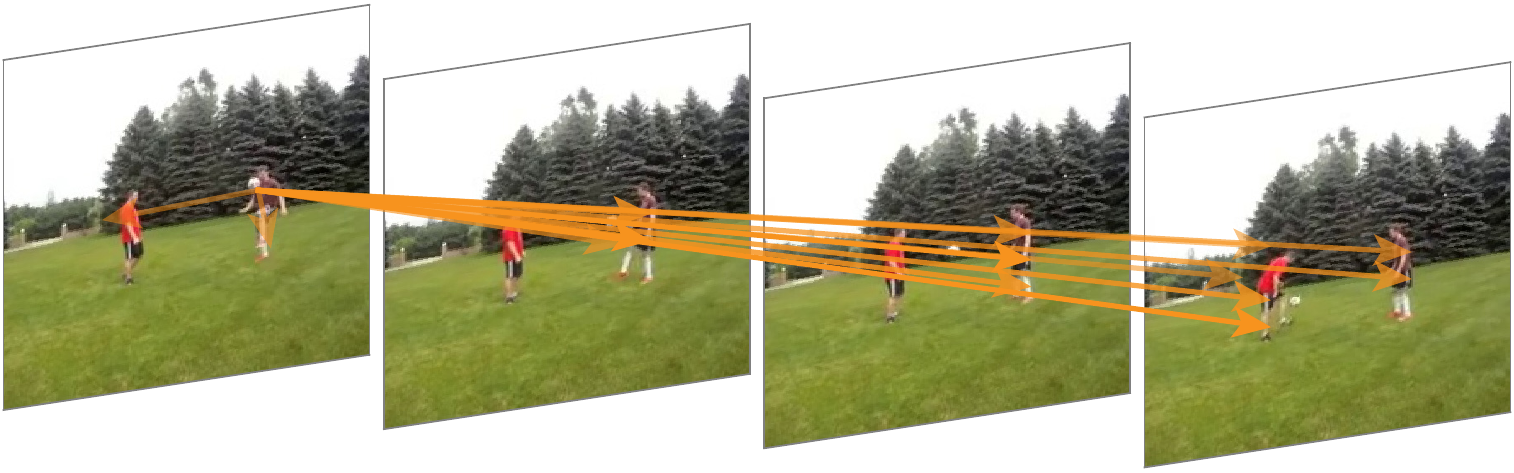}
\includegraphics[width=\sz\linewidth]{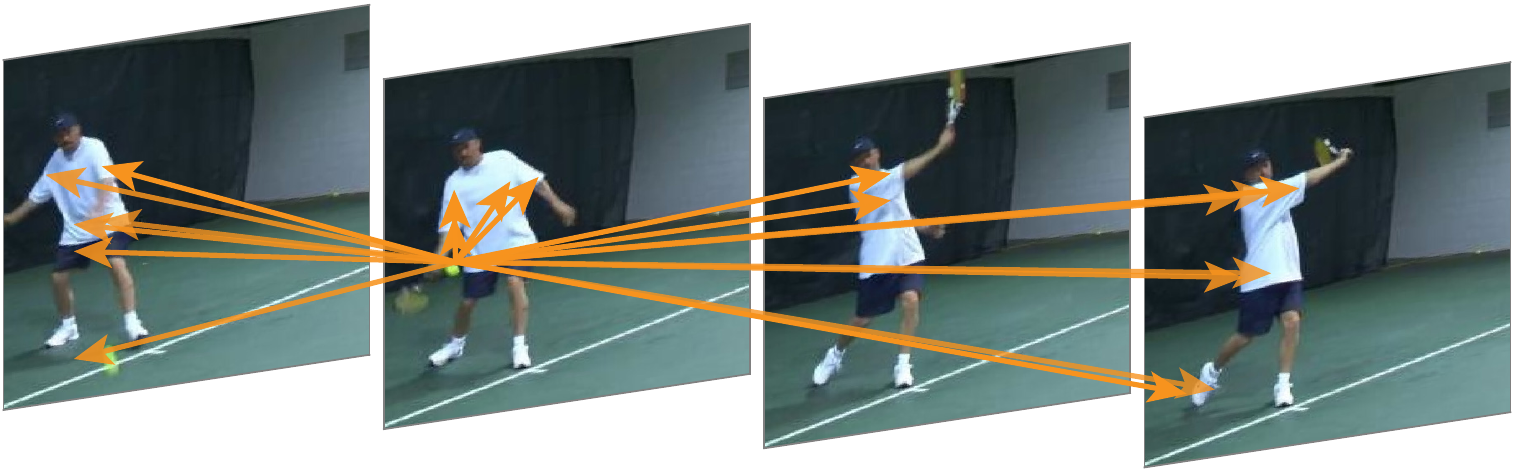}\quad
\includegraphics[width=\sz\linewidth]{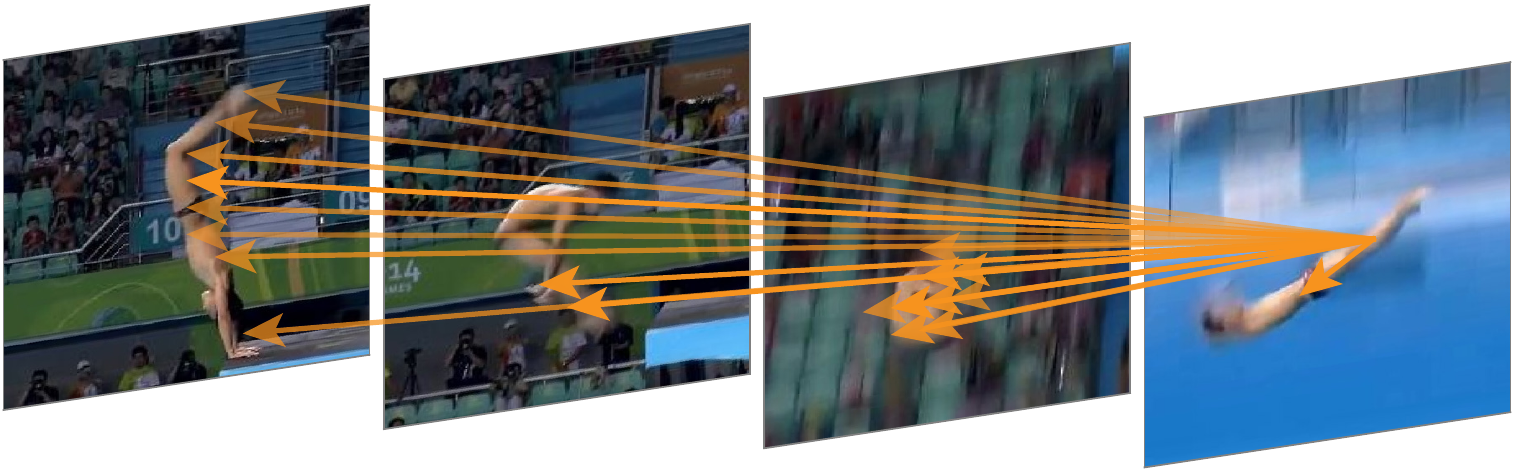}
\caption{Examples of the behavior of a non-local block in res$_3$ computed by a 5-block non-local model trained on Kinetics. These examples are from held-out validation videos.
The starting point of arrows represents one $\ve{x}_i$, and the ending points represent $\ve{x}_j$. The 20 highest weighted arrows for each $\ve{x}_i$ are visualized. The 4 frames are from a 32-frame input, shown with a stride of 8 frames.
These visualizations show how the model finds related clues to support its prediction.}
\label{fig:examples}
\vspace{-.6em}
\end{figure*}
%##################################################################################################

\section{Video Classification Models}

To understand the behavior of non-local networks, we conduct comprehensive ablation experiments on video classification tasks. First we describe our baseline network architectures for this task, and then extend them into 3D ConvNets \cite{Tran2015,Carreira2017} and our proposed non-local nets.

\paragraph{2D ConvNet baseline (C2D).} To isolate the temporal effects of our non-local nets \vs 3D ConvNets, we construct a simple 2D baseline architecture in which the temporal dimension is trivially addressed (\ie, only by pooling).

Table~\ref{tab:arch} shows our C2D baseline under a ResNet-50 backbone. The input video clip has 32 frames each with 224\x224 pixels. All convolutions in Table~\ref{tab:arch} are in essence 2D kernels that process the input frame-by-frame (implemented as 1\x$k$\x$k$ kernels). This model can be directly initialized from the ResNet weights pre-trained on ImageNet. A ResNet-101 counterpart is built in the same way.

The only operation involving the temporal domain are the pooling layers. In other words, this baseline simply aggregates temporal information.

\paragraph{Inflated 3D ConvNet (I3D).} As done in \cite{Feichtenhofer2016,Carreira2017}, one can turn the C2D model in Table~\ref{tab:arch} into a 3D convolutional counterpart by ``inflating'' the kernels. For example, a 2D $k$\x$k$ kernel can be inflated as a 3D $t$\x$k$\x$k$ kernel that spans $t$ frames. This kernel can be initialized from 2D models (pre-trained on ImageNet): each of the $t$ planes in the $t$\x$k$\x$k$ kernel is initialized by the pre-trained $k$\x$k$ weights, rescaled by $1/t$. If a video consists of a single static frame repeated in time, this initialization produces the same results as the 2D pre-trained model run on a static frame.

We study two cases of inflations: we either inflate the 3\x3 kernel in a residual block to 3\x3\x3 (similar to \cite{Carreira2017}), or the first 1\x1 kernel in a residual block to 3\x1\x1 (similar to \cite{Feichtenhofer2016}). We denote these as I3D$_{3\times3\times3}$ and I3D$_{3\times1\times1}$. As 3D convolutions are computationally intensive, we only inflate one kernel for every 2 residual blocks; inflating more layers shows diminishing return.
We inflate conv$_1$ to 5\x7\x7.

The authors of \cite{Carreira2017} have shown that I3D models are more accurate than their CNN+LSTM counterparts.

\paragraph{Non-local network.} We insert non-local blocks into C2D or I3D to turn them into non-local nets. We investigate adding 1, 5, or 10 non-local blocks; the implementation details are described in the next section in context.

\subsection{Implementation Details}

\begin{figure}[t]
\centering
\includegraphics[width=.9\linewidth]{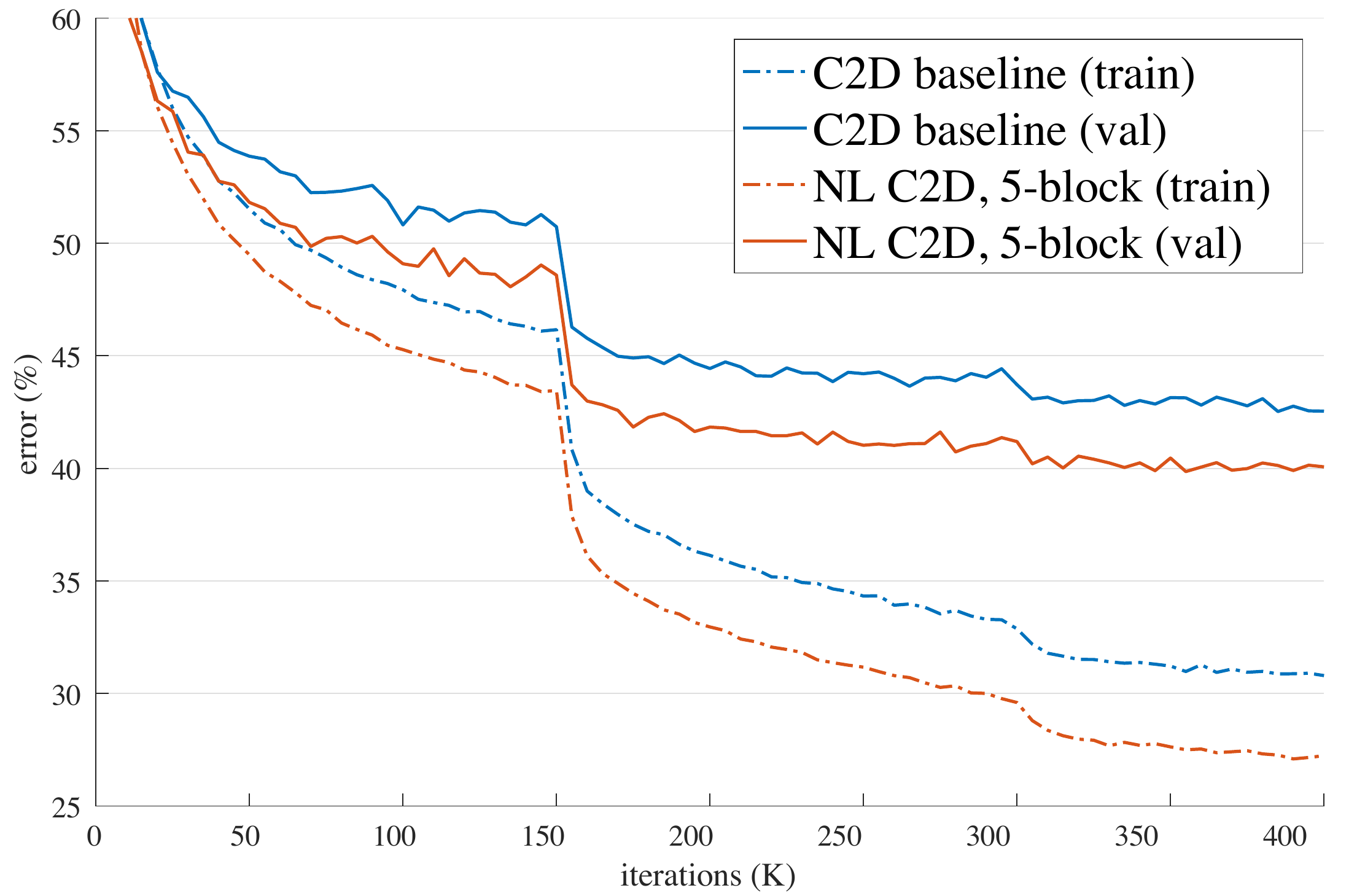}
\caption{Curves of the training procedure on Kinetics for the ResNet-50 C2D baseline (blue) \vs non-local C2D with 5 blocks (red). We show the top-1 training error (dash) and validation error (solid). The validation error is computed in the same way as the training error (so it is 1-clip testing with the same random jittering at training time); the final results are in Table~\ref{tab:ablation:deeper} (R50, 5-block).  }
\label{fig:curves}
\end{figure}

\paragraph{Training.} Our models are pre-trained on ImageNet \cite{Russakovsky2015}. Unless specified, we fine-tune our models using 32-frame input clips. These clips are formed by randomly cropping out 64 consecutive frames from the original full-length video and then dropping every other frame. The spatial size is 224\x224 pixels, randomly cropped from a scaled video whose shorter side is randomly sampled in $[256, 320]$ pixels, following \cite{Simonyan2015}.
We train on an 8-GPU machine and each GPU has 8 clips in a mini-batch (so in total with a mini-batch size of 64 clips). We train our models for 400k iterations in total, starting with a learning rate of 0.01 and reducing it by a factor of 10 at every 150k iterations (see also Figure~\ref{fig:curves}). We use a momentum of 0.9 and a weight decay of 0.0001. We adopt dropout \cite{Hinton2012} after the global pooling layer, with a dropout ratio of 0.5.
We fine-tune our models with BatchNorm (BN) \cite{Ioffe2015} enabled when it is applied. This is in contrast to common practice \cite{He2016} of fine-tuning ResNets, where BN was frozen. We have found that enabling BN in our application reduces overfitting.

We adopt the method in \cite{He2015} to initialize the weight layers introduced in the non-local blocks. We add a BN layer right after the last 1\x1\x1 layer that represents $W_z$; we do not add BN to other layers in a non-local block. The scale parameter of this BN layer is initialized as zero, following \cite{Goyal2017}. This ensures that the initial state of the entire non-local block is an identity mapping, so it can be inserted into any pre-trained networks while maintaining its initial behavior.

\paragraph{Inference.} Following \cite{Simonyan2015} we perform spatially fully-convolutional inference on videos whose shorter side is rescaled to 256.
For the temporal domain, in our practice we sample 10 clips evenly from a full-length video and compute the softmax scores on them individually. The final prediction is the averaged softmax scores of all clips.

%##################################################################################################
\begin{table*}[t]\centering\vspace{-3mm}
% subfloat ############
\subfloat[\textbf{Instantiations}: 1 non-local block of different types is added into the C2D baseline. All entries are with ResNet-50. \label{tab:ablation:instantiations}]{
\tablestyle{3pt}{1.05}
\begin{tabular}{l|x{22}x{22}}
\multicolumn{1}{c|}{model, R50}  & top-1 & top-5 \\
\shline
C2D baseline & 71.8 & 89.7 \\
\hline
Gaussian & 72.5 & 90.2  \\
Gaussian, embed & 72.7 & \bd{90.5} \\
dot-product & \bd{72.9} & 90.3 \\
concatenation & 72.8 & \bd{90.5} \\
 \multicolumn{3}{c}{~}\\
 \multicolumn{3}{c}{~}\\
 \multicolumn{3}{c}{~}\\
\end{tabular}}\hspace{3mm}
% subfloat ############
\subfloat[\textbf{Stages}: 1 non-local block is added into different stages. All entries are with ResNet-50. \label{tab:ablation:stages}]{
\tablestyle{2pt}{1.05}
\begin{tabular}{c|x{22}x{22}}
\multicolumn{1}{c|}{model, R50}  & top-1 & top-5 \\
\shline
baseline & 71.8 & 89.7 \\
\hline
res$_2$ & 72.7 & 90.3 \\
res$_3$ & \bd{72.9} & 90.4 \\
res$_4$ & 72.7 & \bd{90.5} \\
res$_5$ & 72.3 & 90.1  \\
 \multicolumn{3}{c}{~}\\
 \multicolumn{3}{c}{~}\\
 \multicolumn{3}{c}{~}\\
\end{tabular}}\hspace{3mm}
% subfloat ############
\subfloat[\textbf{Deeper non-local models}: we compare 1, 5, and 10 non-local blocks added to the C2D baseline. We show ResNet-50 (top) and ResNet-101 (bottom) results. \label{tab:ablation:deeper}]{
\tablestyle{3pt}{1.05}
\begin{tabular}{lc|x{22}x{22}}
\multicolumn{2}{c|}{model} & top-1 & top-5 \\
\shline
\multirow{4}{*}{R50} & baseline  & 71.8 & 89.7 \\
& 1-block   & 72.7 & 90.5 \\
& 5-block   & 73.8 & 91.0 \\
& 10-block   & \bd{74.3} & \bd{91.2} \\
\hline
\multirow{4}{*}{R101} & baseline  & 73.1 & 91.0 \\
& 1-block   & 74.3 & 91.3 \\
& 5-block   & \bd{75.1} & \bd{91.7} \\
& 10-block   & \bd{75.1} & 91.6 \\
\end{tabular}}\hspace{3mm}
% subfloat ############
\subfloat[\textbf{Space \vs time \vs spacetime}: we compare non-local operations applied along space, time, and spacetime dimensions respectively. 5 non-local blocks are used.  \label{tab:ablation:spacetime}]{
\tablestyle{3pt}{1.05}
\begin{tabular}{ll|x{22}x{22}}
\multicolumn{2}{c|}{model} & top-1 & top-5 \\
\shline
\multirow{4}{*}{R50} & baseline  & 71.8 & 89.7 \\
& space-only & 72.9 & 90.8 \\
& time-only & 73.1 & 90.5 \\
& spacetime & \bd{73.8} & \bd{91.0} \\
\hline
\multirow{4}{*}{R101} & baseline  & 73.1 & 91.0 \\
& space-only & 74.4 & 91.3 \\
& time-only & 74.4 & 90.5 \\
& spacetime & \bd{75.1} & \bd{91.7} \\
\end{tabular}}%\hspace{3mm}

% subfloat ############
\subfloat[\textbf{Non-local \vs 3D Conv}: A 5-block non-local C2D \vs inflated 3D ConvNet (I3D) \cite{Carreira2017}. All entries are with ResNet-101. The numbers of parameters and FLOPs are relative to the C2D baseline (43.2M and 34.2B). \label{tab:ablation:c3d}]{
\tablestyle{3pt}{1.05}
\begin{tabular}{l|x{22}x{22}|x{22}x{22}}
\multicolumn{1}{c|}{model, R101}  & \multicolumn{1}{l}{params} & FLOPs & top-1 & top-5 \\
\shline
C2D baseline  & 1\x & 1\x & 73.1 & 91.0 \\
\hline
I3D$_{3\times3\times3}$ & 1.5\x & 1.8\x & 74.1 &	91.2  \\
I3D$_{3\times1\times1}$ & \bd{1.2\x} & 1.5\x & 74.4 & 91.1 \\
\hline
NL C2D, 5-block & \bd{1.2\x} & \bd{1.2\x} & \bd{75.1} & \bd{91.7} \\
 \multicolumn{3}{c}{~}\\
 \multicolumn{3}{c}{~}\\
\end{tabular}}\hspace{3mm}
\subfloat[\textbf{Non-local 3D ConvNet}: 5 non-local blocks are added on top of our best I3D models. These results show that non-local operations are complementary to 3D convolutions.  \label{tab:ablation:combine}]{
\tablestyle{4pt}{1.05}
\begin{tabular}{ll|x{22}x{22}}
\multicolumn{2}{c|}{model} & top-1 & top-5 \\
\shline
\multirow{3}{*}{R50} & C2D baseline  & 71.8 & 89.7 \\
& I3D & 73.3 & 90.7 \\
& NL I3D & \bd{74.9} & \bd{91.6} \\
\hline
\multirow{3}{*}{R101} & C2D baseline  & 73.1 & 91.0 \\
& I3D & 74.4 & 91.1 \\
& NL I3D & \bd{76.0} & \bd{92.1} \\
\end{tabular}}\hspace{3mm}
% subfloat ############
\subfloat[\textbf{Longer clips}: we fine-tune and test the models in Table~\ref{tab:ablation:combine} on the 128-frame clips. The gains of our non-local operations are consistent.  \label{tab:ablation:long}]{
\tablestyle{4pt}{1.05}
\begin{tabular}{ll|x{22}x{22}}
\multicolumn{2}{c|}{model} & top-1 & top-5 \\
\shline
\multirow{3}{*}{R50} & C2D baseline  & 73.8 & 91.2 \\
%& Non-local & 75.4 & 92.1 \\
& I3D & 74.9 & 91.7 \\
& NL I3D & \bd{76.5} & \bd{92.6} \\
\hline
\multirow{3}{*}{R101} & C2D baseline  & 75.3 & 91.8 \\
%& Non-local & 76.8 & 92.8 \\
& I3D & 76.4 & 92.7 \\
& NL I3D & \bd{77.7} & \bd{93.3} \\
\end{tabular}}%\hspace{3mm}
%\vspace{-.5em}
% main caption
\caption{\textbf{Ablations} on Kinetics action classification. We show top-1 and top-5 classification accuracy (\%).}
\label{tab:ablations}
\end{table*}
%##################################################################################################

%##################################################################################################
%##################################################################################################

\section{Experiments on Video Classification}

We perform comprehensive studies on the challenging Kinetics dataset \cite{Kay2017}. We also report results on the Charades dataset \cite{Sigurdsson2016} to show the generality of our models.

\subsection{Experiments on Kinetics}

Kinetics \cite{Kay2017} contains $\app$246k training videos and 20k validation videos. It is a classification task involving 400 human action categories. We train all models on the training set and test on the validation set.

Figure~\ref{fig:curves} shows the curves of the training procedure of a ResNet-50 C2D baseline \vs a non-local C2D with 5 blocks (more details in the following). Our non-local C2D model is consistently better than the C2D baseline \emph{throughout the training procedure}, in both training and validation error.

Figure~\ref{fig:teaser} and Figure~\ref{fig:examples} visualize several examples of the behavior of a non-local block computed by our models. Our network can learn to find meaningful relational clues regardless of the distance in space and time.

Table~\ref{tab:ablations} shows the ablation results, analyzed as follows:

% ##########
\paragraph{Instantiations.} Table~\ref{tab:ablation:instantiations} compares different types of a single non-local block added to the C2D baseline (right before the last residual block of res$_4$). Even adding one non-local block can lead to  $\app$1\% improvement over the baseline.

Interestingly, the embedded Gaussian, dot-product, and concatenation versions perform similarly, up to some random variations (72.7 to 72.9). As discussed in Sec.~\ref{sec:instantiations}, the non-local operations with Gaussian kernels become similar to the self-attention module \cite{Vaswani2017}. However, our experiments show that the attentional (softmax) behavior of this module is \emph{not} the key to the improvement in our applications; instead, it is more likely that the non-local behavior is important, and it is insensitive to the instantiations.

In the rest of this paper, we use the embedded Gaussian version by default. This version is easier to visualize as its softmax scores are in the range of $[0, 1]$.

% ##########
\paragraph{Which stage to add non-local blocks?} Table~\ref{tab:ablation:stages} compares a single non-local block added to different stages of ResNet. The block is added to right before the last residual block of a stage. The improvement of a non-local block on res$_2$, res$_3$, or res$_4$ is similar, and on res$_5$ is slightly smaller. One possible explanation is that res$_5$ has a small spatial size (7\x7) and it is insufficient to provide precise spatial information. More evidence of a non-local block exploiting spatial information will be investigated in Table~\ref{tab:ablation:spacetime}.

%##################################################################################################
\definecolor{demphcolor}{RGB}{144,144,144}
\newcommand{\demph}[1]{\textcolor{demphcolor}{#1}}
\begin{table*}[h!]
\centering
\small
\tablestyle{4pt}{1.05}
\begin{tabular}{l|l|l|x{30}x{30}|x{30}x{30}x{30}}
\multicolumn{1}{c|}{model} & \multicolumn{1}{c|}{backbone} & \multicolumn{1}{c|}{modality}  & top-1 val & top-5 val & top-1 test & top-5 test & avg test$^\dagger$  \\
\shline
I3D in \cite{Carreira2017} & Inception & RGB & 72.1 & 90.3 & 71.1 & 89.3 & 80.2 \\
2-Stream I3D in \cite{Carreira2017} &  Inception & RGB + flow & 75.7 & 92.0 & 74.2 & 91.3 & 82.8 \\
\hline
RGB baseline in \cite{Bian2017} & Inception-ResNet-v2 & RGB & 73.0 & 90.9 & - & - & - \\
\demph{3-stream late fusion \cite{Bian2017}} & \demph{Inception-ResNet-v2} & \demph{RGB + flow + audio} & \demph{74.9} & \demph{91.6} & \demph{-} & \demph{-} & \demph{-} \\
\demph{3-stream LSTM \cite{Bian2017}} & \demph{Inception-ResNet-v2}  & \demph{RGB + flow + audio}  & \demph{77.1} & \demph{93.2} & \demph{-} & \demph{-} & \demph{-} \\
\demph{3-stream SATT \cite{Bian2017}} & \demph{Inception-ResNet-v2}  & \demph{RGB + flow + audio}  & \demph{77.7} & \demph{93.2} & \demph{-} & \demph{-} & \demph{-} \\
\hline
\multirow{2}{*}{NL I3D [ours]}  & ResNet-50 & RGB & 76.5 & 92.6 & - & - & - \\
& ResNet-101 & RGB & \bd{77.7} & \bd{93.3} & - & - & \bd{83.8} \\
\end{tabular}
\vspace{.2em}
\caption{Comparisons with state-of-the-art results in \textbf{Kinetics}, reported on the val and test sets.
We include the Kinetics 2017 competition winner's results \cite{Bian2017}, but their best results exploited audio signals (marked in \demph{gray}) so were not vision-only solutions. $^\dagger$: ``avg'' is the average of top-1 and top-5 accuracy;  individual top-1 or top-5 numbers are not available from the test server at the time of submitting this manuscript.}
\label{tab:kinetics}
\vspace{-.7em}
\end{table*}
%##################################################################################################

% ##########
\paragraph{Going deeper with non-local blocks.} Table~\ref{tab:ablation:deeper} shows the results of more non-local blocks. We add 1 block (to res$_4$), 5 blocks (3 to res$_4$ and 2 to res$_3$, to every other residual block), and 10 blocks (to every residual block in res$_3$ and res$_4$) in ResNet-50; in ResNet-101 we add them to the corresponding residual blocks. Table~\ref{tab:ablation:deeper} shows that more non-local blocks in general lead to better results.
We argue that multiple non-local blocks can perform long-range multi-hop communication. Messages can be delivered back and forth between distant positions in spacetime, which is hard to do via local models.

It is noteworthy that the improvement of non-local blocks is \emph{not} just because they add depth to the baseline model. To see this, we note that in Table~\ref{tab:ablation:deeper} the non-local 5-block ResNet-50 model has 73.8 accuracy, higher than the deeper ResNet-101 baseline's 73.1. However, the 5-block ResNet-50 has only $\app$70\% parameters and $\app$80\% FLOPs of the ResNet-101 baseline, and is also \emph{shallower}. This comparison shows that the improvement due to non-local blocks is complementary to going deeper in standard ways.

We have also tried to add standard residual blocks, instead of non-local blocks, to the baseline models. The accuracy is not increased. This again shows that the improvement of non-local blocks is not just because they add depth.

% ##########
\paragraph{Non-local in spacetime.} Our method can naturally handle spacetime signals. This is a nice property: related objects in a video can present at distant space and long-term time interval, and their dependency can be captured by our model.

In Table~\ref{tab:ablation:spacetime} we study the effect of non-local blocks applied along space, time, or spacetime. For example, in the space-only version, the non-local dependency only happens within the same frame: \ie, in Eq.(\ref{eq:nonlocal}) it only sums over the index $j$ in the same frame of the index $i$. The time-only version can be set up similarly. Table~\ref{tab:ablation:spacetime} shows that both the space-only and time-only versions improve over the C2D baseline, but are inferior to the spacetime version.

% ##########
\paragraph{Non-local net \vs 3D ConvNet.} Table~\ref{tab:ablation:c3d} compares our non-local C2D version with the inflated 3D ConvNets. Non-local operations and 3D convolutions can be seen as two ways of extending C2D to the temporal dimensions.

Table~\ref{tab:ablation:c3d} also compares the number of parameters and FLOPs, relative to the baseline. Our non-local C2D model is more accurate than the I3D counterpart (\eg, 75.1 \vs 74.4), while having a smaller number of FLOPs (1.2\x~\vs 1.5\x). This comparison shows that our method can be more effective than 3D convolutions when used alone.

% ##########
\paragraph{Non-local 3D ConvNet.} Despite the above comparison, non-local operations and 3D convolutions can model different aspects of the problem: 3D convolutions can capture local dependency.  Table~\ref{tab:ablation:combine} shows the results of inserting 5 non-local blocks into the I3D$_{3\times1\times1}$ models. These non-local I3D (NL I3D) models improve over their I3D counterparts (+1.6 point accuracy), showing that non-local operations and 3D convolutions are complementary.

% ##########
\paragraph{Longer sequences.} Finally we investigate the generality of our models on longer input videos. We use input clips consisting of 128 consecutive frames without subsampling.
The sequences throughout all layers in the networks are thus 4\x~longer compared to the 32-frame counterparts. To fit this model into memory, we reduce the mini-batch size to 2 clips per GPU. As a result of using small mini-batches, we freeze all BN layers in this case. We initialize this model from the corresponding models trained with 32-frame inputs. We fine-tune on 128-frame inputs using the same number of iterations as the 32-frame case (though the mini-batch size is now smaller), starting with a learning rate of 0.0025. Other implementation details are the same as before.

Table~\ref{tab:ablation:long} shows the results of 128-frame clips. Comparing with the 32-frame counterparts in Table~\ref{tab:ablation:combine}, all models have better results on longer inputs. We also find that our NL I3D can maintain its gain over the I3D counterparts, showing that our models work well on longer sequences.

% ##########
\paragraph{Comparisons with state-of-the-art results.} Table~\ref{tab:kinetics} shows the results from the I3D authors \cite{Carreira2017} and from the Kinetics 2017 competition winner \cite{Bian2017}. We note that these are comparisons of systems which can differ in many aspects. Nevertheless, our method surpasses all the existing RGB or RGB + flow based methods by a good margin. \emph{Without using optical flow and without any bells and whistles}, our method is on par with the heavily engineered results of the 2017 competition winner.

\begin{table}[t]
\centering
\small
\tablestyle{6pt}{1.05}
\begin{tabular}{l|l|x{36}x{36}}
\multicolumn{1}{c|}{model}  & \multicolumn{1}{c|}{modality}  & \emph{train/val} & \emph{trainval/test} \\
\shline
2-Stream~\cite{Sigurdsson2017} & RGB + flow & 18.6 & - \\
2-Stream +LSTM~\cite{Sigurdsson2017} & RGB + flow & 17.8 & - \\
Asyn-TF~\cite{Sigurdsson2017} & RGB + flow & 22.4 & - \\
I3D~\cite{Carreira2017} & RGB & 32.9 & 34.4 \\
\hline
I3D [ours]   & RGB & 35.5 &  37.2  \\
NL I3D [ours] & RGB & \bd{37.5} & \bd{39.5} \\
\end{tabular}
\vspace{0.5em}
\caption{Classification mAP (\%) in the \textbf{Charades} dataset \cite{Sigurdsson2016}, on the \emph{train/val} split and the \emph{trainval/test} split. Our results are based on ResNet-101. Our NL I3D uses 5 non-local blocks. \vspace{-1em}}
\label{tab:charades}
\end{table}

\vspace{-0.5em}
\subsection{Experiments on Charades}
\vspace{-0.2em}
Charades \cite{Sigurdsson2016} is a video dataset with $\app$8k training, $\app$1.8k validation, and $\app$2k testing videos. It is a multi-label classification task with 157 action categories. We use a per-category sigmoid output to handle the multi-label property.

We initialize our models pre-trained on Kinetics (128-frame).
The mini-batch size is set to 1 clip per GPU. We train our models for 200k iterations, starting from a learning rate of 0.00125 and reducing it by 10 every 75k iterations.
We use a jittering strategy similar to that in Kinetics to determine the location of the 224\x224 cropping window, but we rescale the video such that this cropping window outputs 288\x288 pixels, on which we fine-tune our network.
We test on a single scale of 320 pixels.

Table~\ref{tab:charades} shows the comparisons with the previous results on Charades. The result of \cite{Carreira2017} is the 2017 competition winner in Charades, which was also fine-tuned from models pre-trained in Kinetics. Our I3D baseline is higher than previous results. As a controlled comparison, our non-local net improves over our I3D baseline by 2.3\% on the test set.

\vspace{-0.5em}
\section{Extension: Experiments on COCO}
\vspace{-0.5em}
We also investigate our models on static image recognition.
We experiment on the Mask R-CNN baseline \cite{He2017} for COCO \cite{Lin2014} object detection/segmentation and human pose estimation (keypoint detection). The models are trained on COCO \texttt{train2017} (\ie, \texttt{trainval35k} in 2014) and tested on \texttt{val2017} (\ie, \texttt{minival} in 2014).

\paragraph{Object detection and instance segmentation.}

We modify the Mask R-CNN backbone by adding one non-local block (right before the last residual block of res$_4$). All models are fine-tuned from ImageNet pre-training. We evaluate on a standard baseline of ResNet-50/101 and a high baseline of ResNeXt-152 (X152) \cite{Xie2017}. Unlike the original paper \cite{He2017} that adopted stage-wise training regarding RPN, we use an improved implementation with end-to-end joint training similar to \cite{Ren2017}, which leads to higher baselines than \cite{He2017}.

Table~\ref{tab:coco_det} shows the box and mask AP on COCO. We see that a single non-local block improves all R50/101 and X152 baselines, on all metrics involving detection and segmentation. AP$^\text{box}$ is increased by $\app$1 point in all cases (\eg, +1.3 point in R101). Our non-local block is \emph{complementary} to increasing the model capacity, even when the model is upgraded from R50/101 to X152. This comparison suggests that \emph{non-local dependency has not been sufficiently captured by existing models despite increased depth/capacity}.

In addition, the above gain is at a very small cost. The single non-local block only adds $<$5\% computation to the baseline model. We also have tried to use more non-local blocks to the backbone, but found diminishing return.

\begin{table}[t]
\centering
\small
\tablestyle{3pt}{1.05}\begin{tabular}{cc|x{22}x{22}x{22}|x{22}x{22}x{22}}
 \multicolumn{2}{c|}{method} & AP$^\text{box}$ & AP$^\text{box}_{50}$ & AP$^\text{box}_{75}$
 & AP$^\text{mask}$ & AP$^\text{mask}_{50}$ & AP$^\text{mask}_{75}$ \\[.1em]
\shline
\multirow{2}{*}{R50} & baseline & 38.0 & 59.6 & 41.0 & 34.6 & 56.4 & 36.5 \\
& +1 NL & \bd{39.0} & \bd{61.1} & \bd{41.9} & \bd{35.5} & \bd{58.0} & \bd{37.4}  \\
\hline
\multirow{2}{*}{R101} & baseline & 39.5 & 61.4 & 42.9 & 36.0 & 58.1 & 38.3 \\
& +1 NL & \bd{40.8} & \bd{63.1} & \bd{44.5} & \bd{37.1} & \bd{59.9} & \bd{39.2}  \\
\hline
\multirow{2}{*}{X152} & baseline & 44.1 & 66.4 & 48.4 & 39.7 & 63.2 & 42.2 \\
& +1 NL & \bd{45.0} & \bd{67.8} & \bd{48.9} & \bd{40.3} & \bd{64.4} & \bd{42.8} \\
\end{tabular}
\vspace{0.5em}
\caption{Adding 1 non-local block to Mask R-CNN for COCO \bd{object detection} and \bd{instance segmentation}. The backbone is ResNet-50/101 or ResNeXt-152 \cite{Xie2017}, both with FPN \cite{Lin2017}.}
\label{tab:coco_det}
\vspace{-0.5em}
\end{table}

\paragraph{Keypoint detection.}

Next we evaluate non-local blocks in Mask R-CNN for keypoint detection. In \cite{He2017}, Mask R-CNN used a stack of 8 convolutional layers for predicting the keypoints as 1-hot masks. These layers are local operations and may overlook the dependency among keypoints across long distance. Motivated by this, we insert 4 non-local blocks into the keypoint head (after every 2 convolutional layers).

Table~\ref{tab:coco_kp} shows the results on COCO. On a strong baseline of R101, adding 4 non-local blocks to the keypoint head leads to a $\app$1 point increase of keypoint AP. If we add one extra non-local block to the backbone as done for object detection, we observe an in total 1.4 points increase of keypoint AP over the baseline. In particular, we see that the stricter criterion of AP$_{75}$ is boosted by 2.4 points, suggesting a stronger localization performance.

\begin{table}[t]
\centering
\small
\tablestyle{4pt}{1.05}
\begin{tabular}{l|x{22}x{22}x{22}}
 model & AP$^\text{kp}$ & AP$^\text{kp}_{50}$ & AP$^\text{kp}_{75}$\\ [.1em]
\shline
R101 baseline  & 65.1 & 86.8 & 70.4 \\
\hline
NL, +4 in head & 66.0 & 87.1 & 71.7 \\
NL, +4 in head, +1 in backbone  & \bd{66.5} & \bd{87.3} & \bd{72.8}
\end{tabular}
\vspace{0.5em}
\caption{Adding non-local blocks to Mask R-CNN for COCO \bd{keypoint detection}. The backbone is ResNet-101 with FPN \cite{Lin2017}.\vspace{-1em}}
\label{tab:coco_kp}
\end{table}

\vspace{-0.5em}
\section{Conclusion}
\vspace{-0.5em}
We presented a new class of neural networks which capture long-range dependencies via non-local operations. Our non-local blocks can be combined with any existing architectures. We show the significance of non-local modeling for the tasks of video classification, object detection and segmentation, and pose estimation. On all tasks, a simple addition of non-local blocks provides solid improvement over baselines. We hope non-local layers will become an important component of future network architectures.

{\footnotesize
{\noindent {\bf Acknowledgement}: This work was partially supported by ONR MURI N000141612007, Sloan, Okawa Fellowship to AG and NVIDIA Fellowship to XW. We would also like to thank Haoqi Fan, Du Tran, Heng Wang, Georgia Gkioxari and Piotr Dollar for many helpful discussions.}
}

{
\small
\bibliographystyle{ieee}
\bibliography{nonlocal}
}

\end{document}